\documentclass{article}

\usepackage{arxiv}

\usepackage[utf8]{inputenc} % allow utf-8 input

\usepackage{amsfonts}       % blackboard math symbols
\usepackage{nicefrac}       % compact symbols for 1/2, etc.
\usepackage{microtype}      % microtypography
\usepackage{lipsum}

\usepackage{color}
\usepackage{listings}
\usepackage{booktabs}
\usepackage{array}
\usepackage{graphicx}
\usepackage{longtable}
\usepackage{subfigure}

\usepackage{cite}
\usepackage{url}
\usepackage{fancyhdr}

\usepackage{soul}

\usepackage{float}
\usepackage{listings}
\usepackage{mdwmath}
\usepackage{mdwtab}
\usepackage{multirow}
\usepackage{multicol}
\usepackage{rotating}
\usepackage{setspace}
\usepackage[utf8]{inputenc}
\usepackage{lineno}
\usepackage{listings}

\usepackage{mdwmath}
\usepackage{mdwtab}
\usepackage{multirow}
\usepackage{multicol}
\usepackage{array}
\usepackage{booktabs}

\usepackage{enumitem}
\usepackage{xspace}
\usepackage[export]{adjustbox}
\usepackage{graphicx}
\usepackage{color,soul}
\usepackage{rotating}
\usepackage{setspace}
\usepackage{amsmath} 
\usepackage{amssymb}
\usepackage{float}
\usepackage[dvipsnames]{xcolor}

\usepackage{hyperref}
\usepackage{url}

\usepackage{multirow}
\usepackage{booktabs}
\usepackage[normalem]{ulem}
\useunder{\uline}{\ul}{}

\newcommand{\eg}{\textit{e}.\textit{g}.,\xspace}

\usepackage{array}
\usepackage{longtable}
\usepackage{geometry}
\newcolumntype{L}[1]{>{\raggedright\let\newline\\\arraybackslash\hspace{0pt}}m{#1}}
\newcolumntype{C}[1]{>{\centering\let\newline\\\arraybackslash\hspace{0pt}}m{#1}}
\newcolumntype{R}[1]{>{\raggedleft\let\newline\\\arraybackslash\hspace{0pt}}m{#1}}

\usepackage{forest}
\usepackage{caption}

\usepackage{tikz}
\usetikzlibrary{arrows}
\usetikzlibrary{trees}

\title{AgentOps: Enabling Observability of LLM Agents}
\author{Liming Dong, Qinghua Lu, Liming Zhu \\
Data61, CSIRO, Australia
}
\date{Nov, 2024}

\begin{document}

\maketitle

\begin{abstract}

%The ever-improving quality of LLMs has fueled the growth of a diverse range of downstream tasks, leading to an increased demand for AI automation and a burgeoning interest in developing foundation model (FM)-based autonomous agents. 
%As AI agent systems tackle more complex tasks and evolve, they involve a wider range of stakeholders, including agent users, agentic system developers and deployers, and AI model developers. These systems also integrate multiple components such as AI agent workflows, RAG pipelines, prompt management, agent capabilities, and observability features.
%In this case, obtaining reliable outputs and answers from these agents remains challenging, necessitating a dependable execution process and end-to-end observability solutions. 
%To build reliable AI agents and LLM applications, it is essential to shift towards designing AgentOps/LLMOps platforms that ensure observability and traceability across the entire development-to-production life-cycle.
%To this end, we conducted a rapid review and identified relevant AgentOps tools from the agentic ecosystem. 
%Based on this review, we provide an overview of the essential features of AgentOps and propose a comprehensive overview of observability data/traceable artifacts across the agent production life-cycle.
%Our findings provide a systematic overview of the current AgentOps landscape, emphasizing the critical role of observability/traceability in enhancing the reliability of autonomous agent systems.

Large language model (LLM) agents have demonstrated remarkable capabilities across various domains, gaining extensive attention from academia and industry. However, these agents raise significant concerns on AI safety due to their autonomous and non-deterministic behavior, as well as continuous evolving nature . From a DevOps perspective, enabling observability in agents is necessary to ensuring AI safety, as stakeholders can gain insights into the agents' inner workings, allowing them to proactively understand the agents, detect anomalies, and prevent potential failures. Therefore, in this paper, we present a comprehensive taxonomy of AgentOps, identifying the artifacts and associated data that should be traced throughout the entire lifecycle of agents to achieve effective observability. The taxonomy is developed based on a systematic mapping study of existing AgentOps tools. Our taxonomy serves as a reference template for developers to design and implement AgentOps infrastructure that supports monitoring, logging, and analytics. thereby ensuring AI safety.
\end{abstract}

\section{Introduction}

An large language model (LLM) is a large-scale language model with tens of billions of parameters, pretrained on vast and diverse datasets, and applicable to various downstream tasks~\cite{bommasani2021opportunities}. 
While LLMs demonstrate impressive capabilities, they also exhibit limitations in understanding and performing complex tasks. This has led to an increasing demand for LLM agents (including agentic systems which are designed to prompt an LLM multiple times using agent-like design patterns and exhibit varying degrees of agent-like behavior), which have the ability to handle complex tasks autonomously~\cite{liu2024agentdesignpatterncatalogue}. 

An LLM agent is an autonomous system powered by LLMs, capable of perceiving context, reasoning, planning,  executing workflows, by leveraging external tools, knowledge bases and other agents to achieve human goals~\cite{lu2024towards}. 
LLM agents have shown remarkable potential to enhance productivity across various domains, attracting widespread attention from academia and industry. For example, many agents are being successfully applied in the software engineering domain, such as Devin \footnote{Devin,  \url{https://www.cognition.ai/blog/introducing-devin}}, ChatDev \footnote{ChatDev. \url{https://github.com/OpenBMB/ChatDev}}, SWE-agent \footnote{SWE-agent, \url{https://github.com/princeton-nlp/SWE-agent}}. Hereafter, we use "agents" to refer specifically to LLM agents throughout this paper.

Despite the huge potential to enhance productivity, the adoption of LLM agents introduces unique challenges due to their inherent characteristics. 
\begin{itemize}
    \item \textbf{Complex Artefacts and Pipelines}: The agents are compound AI systems, integrating LLMs with various components (i.e., design time artifacts), such as context engine and external tools, and dynamically generating runtime artifacts, such as goals and plans. The operational pipelines typically include context processing, reasoning and planning, workflow execution, and continuous evolution based on the feedback. Throughout these processes, the pipelines may leverage external tools, knowledge bases, and other agents to achieve human goals.
    \item \textbf{Autonomy}: The agents operate with a high degree of autonomy, dynamically interact with external environments, including shifting context, external knowledge bases and tools. These interactions are not predetermined, which may increase the risk of unintended behaviors (such as selecting an external tool with vulnerability issues) and pose severe AI safety challenges~\cite{lu2023responsible}. 
    \item \textbf{Non-Deterministic Behaviour}: Due to the probabilistic nature of LLMs, the agents often exhibit non-deterministic behaviour, producing varied outputs even when given the same inputs. This lack of repeatability may lead to unintended consequences, making it challenging to ensure consistent and predictable outcomes.
    \item \textbf{Continuous Evolution}: The agents can evolve over time through continuous learning, adapting based on runtime evaluation results or human feedback. While this adaptability enhance the agents' capabilities and skills, it also introduces further challenges in maintaining alignment with intended quality and safety goals over the agent's lifecycle.
    \item \textbf{Shared Accountability}: The responsibility of the behaviour or decisions of the agents is often shared among multiple stakeholders, including the agent owner, the FM provider, and various providers of external tools/agents. This complicates the identification of failure sources and the assignment of accountability in the event of incidents.
\end{itemize}

To address the challenges outlined above, it is essential for DevOps tools to support observability features, enabling stakeholders to monitor agent behaviour, track the status of artifacts, log associated data, detect anomalies, trace the evolution of artifacts, and assign accountability if incidents occur. \textbf{Observability} refers to the ability to gain actionable insights into the inner workings of an agent by analysing the inputs and outputs (i.e., runtime artifacts) of different components (i.e., design time artifacts), as they flow through the operational pipelines~\cite{bass25engineering}. However, most existing DevOps tools for agents primarily focus on LLM-specific metrics and prompt management, with limited support for the observability of agent-specific artifacts such as goals, plans, and tools. This limitation results in insufficient observability from both the system and pipeline perspectives.

To bridge this gap, we propose \textbf{AgentOps}, a specialised DevOps paradigm tailored for agents. AgentOps provides a holistic view of agent operations, enabling comprehensive observability by systematically tracing of agent artifacts and associated data. In this paper, we perform a systematic mapping study on the existing DevOps tools for monitoring agents and/or agentic systems to understand their features and limitations. Based on the study results, we first propose an artifact relationship model to identify key agent artifacts and their relationship. Then we present a comprehensive taxonomy of AgentOps, detailing the artifacts and associated data that should be traced. Our taxonomy serves as a reference template for developers to design and implement AgentOps tools to support monitoring, logging, and analytics of agents.

The remainder of the paper is organized as follows. Section \ref{sec:methodology} introduces the methodology. 
Section \ref{sec:Mapping Study Result} presents the mapping study results of AgentOps tools and core features. 
Section \ref{sec:Taxonomy} identifies the agent artifacts and their relationship and presents the taxonomy of AgentOps. Section \ref{sec:ttv} addresses threats to validity, and Section \ref{sec:conlusion} concludes the paper and outlines future work.

\section{Methodology} \label{sec:methodology}
\begin{figure*}[ht]
\centering
 \includegraphics[width=0.8\textwidth]{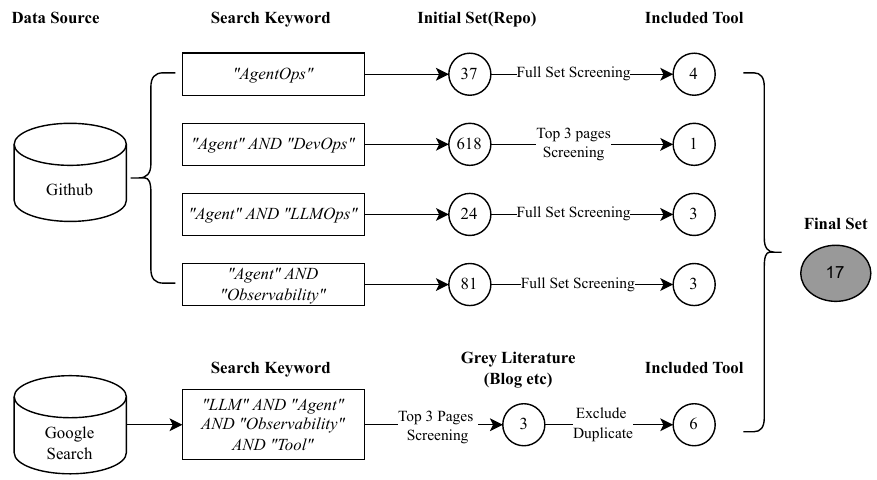}
\caption{Search Process of AgentOps Relevant Tools} \label{fig:Search Process}
\end{figure*}

In this section, we introduce the methodology for this study. Figure\ref{fig:Search Process} provides the overview of the search process. Following the guidelines outlined in \cite{keele2007guidelines, MultiSLR2019},
we performed a systematic mapping study to analyse the existing tools related to AgentOps. This study aimed to understand their features and limitations and to identify the agent artifacts and associated data that should be traced to enable observability.

\vspace{-0.3cm}

\subsection{Data Sources}
\label{subsec:data source}
To identify tools relevant to AgentOps, we used multiple sources. Initially, we conducted a comprehensive search on GitHub\footnote{Github.\url{https://github.com}} using carefully selected keywords. The repositories were then filtered based on predefined selection criteria to ensure relevance and quality. To complement the GitHub search and address potential gaps, we performed a targeted Google Search\footnote{Google Search.\url{https://google.com}} to capture additional tools that may not have been available or visible in the GitHub search results. This multi-source approach ensured broad coverage of both open-source and proprietary AgentOps relevant tools.

\subsection{Search String} 
\label{subsec:search string}
The search terms we used in GitHub primarily focused on the keywords \emph{((“AgentOps”) OR (“Agent” AND “DevOps”) OR (“Agent” AND “LLMOps”))}. 
Since the concept of AgentOps and the exploration of related tools are still in the early stages, we extended the scope of our search to include observability tools that provide comprehensive visibility into LLM applications, must including agents tracing and observability features as well. As a result, we also incorporated the search term \emph{(“Agent” AND “Observability”)}. The search keyword term used in Google Search is \emph{(“LLM” AND “Agent” AND “Observability” AND “Tool”)}.

\begin{table*}[h]
\center
\footnotesize
\caption{Selection Criteria}
\label{tab:AgentOps Selection Criteria}
\begin{tabular}
{@{\extracolsep\fill}L{5mm}|L{100mm}}  
\toprule \hline
\textbf{ID} & \textbf{Inclusion Criteria}\\
\hline
$I1$ & Support observability features (such as monitoring and tracing).\\
$I2$ & Support agent specific tracing or LLM application tracing that can be applied to agents, not just LLM-level tracing. \\ 
$I3$ & Formal release version. \\ 
$I4$ & Public online document available. \\ \hline

\bottomrule

\end{tabular}
\end{table*}

\subsection{Selection Criteria} \label{subsec:selection criteria}
We conducted the tool selection process from the multi-channel data sources mentioned above. 
For this paper, we focused on AgentOps relevant tools that met the criteria illustrated in Table 1.
To ensure robust tracking and management of agent performance and interactions, it is essential that selected tools support core observability functionalities ($I1$) like monitoring and tracing etc. Furthermore, given that the focus of our study is on AgentOps rather than general LLMOps or model-level operations, the tools must support tracing at the agent level or LLM application level ($I2$) that allows us to observe a finer-grained understanding of agent-specific behaviors.
To facilitate both implementation and community-wide adoption, it is essential that tools with formal release version ($I3$) and have accessible and public online documentation ($I4$).

\subsection{Search Process}

%In this study, we systematically identified relevant tools for AgentOps with observability features. The data sources included GitHub for open-source repositories and Google Search for supplementary gray literature. Using GitHub as the primary data source, we applied four specific search keywords to capture a broad range of tools within the AgentOps ecosystem: \emph{((“AgentOps”) OR (“Agent” AND “Observability”) OR (“Agent” AND “LLMOps”) OR (“Agent” AND “DevOps”))}.  

We began by querying GitHub with four specific keyword strings \emph{“AgentOps”}, \emph{“Agent” AND “DevOps”}, \emph{“Agent” AND “LLMOps”}, and \emph{“Agent” AND “Observability”}.
Each query returned an initial set of repositories. For the search strings \emph{(“AgentOps”) OR (“Agent” AND “LLMOps”) OR (“Agent” AND “Observability”)}, we obtained 37, 24, and 81 initial repositories, respectively. All repositories from these search results were screened to ensure relevance, leading to the inclusion of 4, 3, and 3 tools for further analysis, respectively.

For the \emph{(“Agent” AND “DevOps”)} query, the search yielded a much larger set of 618 repositories. To manage this, we limited our screening to the top three pages of results, sorting repositories based on relevance and popularity. Most of these projects, however, were focused on agents for DevOps tasks rather than DevOps platforms for managing agents. As a result, only one additional tool was included from this query.

To ensure comprehensive coverage and address potential gaps in the GitHub search, we conducted an additional search using Google Search to capture proprietary tools. We used the keyword string \emph{(“LLM” AND “Agent” AND “Observability” AND “Tool”)} and screened the top three pages of results. This search targeted relevant blogs, product announcements, and other online resources. After removing duplicates and irrelevant results, this process identified six additional tools.

The combined search process across GitHub and Google Search resulted in a total of 17 tools. This final set formed the basis for our systematic analysis of AgentOps relevant tools' features and limitations.

\subsection{Data Extraction} \label{subsec:DE}
To better understand the current landscape of AgentOps tools, we systematically categorized and summarized 17 selected tools. Relevant data items were extracted from each tool's GitHub repository, official product website, and available project documentation. The specific data items extracted are outlined in Table \ref{tab:Data Extraction}.

\begin{table*}[h]
\center
\footnotesize
\caption{Data Extraction}
\label{tab:Data Extraction}
\begin{tabular}
{@{\extracolsep\fill}L{5mm}|L{30mm}|L{120mm}}  
\toprule \hline
\textbf{ID}&\textbf{Data Item} & \textbf{Description}\\
\hline
$D1$&Name & The name of AgentOps relevant tool. \\ \hline
$D2$&Source & The data source used to identify the included tool. \\\hline
$D3$&GitHub Repo URL & The Github repository URL of included tool.\\\hline
$D4$&Star & The number of stars received on Github (as a proxy for popularity).\\ \hline
$D5$&Scope & The scope of the tool, specifically whether it is designed for tracing agents or LLM applications.\\ \hline
$D6$&Key Features & The primary features provided by the tool for AgentOps or LLM application observability.\\\hline
$D7$&Traceable Artifacts & The artifacts and associated data that the tool tracks. \\ \hline

\bottomrule

\end{tabular}
\end{table*}

\section{Mapping Study Result} \label{sec:Mapping Study Result}

%\textcolor{blue}{
%(Liming Dong) In progress... Explain Mapping Relationship for Tool and Key Features in Table \ref{tab:Key Features Mapping} }

In this section, we present the results of the systematic mapping study on tools relevant to AgentOps.
\begin{table*}[h]
\center
\footnotesize
\caption{List of Tools relevant to AgentOps}
\label{tab:Tool List}
\begin{tabular}
{@{\extracolsep\fill}L{18mm}|L{15mm}|L{38mm}|L{10mm}|L{25mm}}  
\toprule \hline
\textbf{Name}	&\textbf{Source}&	\textbf{Github Repo URL}&	\textbf{Star}&	\textbf{Scope} \\ \hline

Agenta                                & Github        &  Agenta-AI/agenta                     & 1.3k                          &    LLM applications                                                                                                    \\ \hline
AgentNeo                              & Github        & raga-ai-hubAgentNeo& 1k                            & Agents                                                                                                    \\ \hline
AgentOps& Github & AgentOps-AI/agentops & 2.1k & Agents                                                                                               \\ \hline
AGIFlow                               & Github        & AgiFlow/agiflow-sdks                 & 21                            & Agents                                                                                                  \\ \hline
Arize                                 & Google & Arize-Phoenix                      & 3.9k                          &       LLM applications                                                                                                \\ \hline
DataDog                               & Github        & DataDog/datadog-agent               & 2.9k                          &    Agents                                                                                             \\ \hline
Dify                                  & Github        &  langgenius/dify                     & 51.6k                         &        LLM applications                                                                                           \\ \hline
Helicone & Github        & Helicone/helicone                                                  & 1.9k                          &   LLM applications                                                                                                    \\ \hline
Laminar                               & Github        & lmnr-ai/lmnr                          & 1.1k                          &    LLM applications                                                                                                   \\ \hline
Langfuse                              & Github        & langfuse/langfuse                                                  & 6.5k                          &   LLM applications                                                                                                    \\ \hline
LangSmith                             & Github        &  langchain-ai/langsmith-sdk                                                                   & 417                           &     LLM applications                                                                                                  \\ \hline

LangTrace                             & Github        & Scale3-Labs/langtrace               & 552                           &    LLM applications                                                                                                    \\ \hline

Lunary                                & Google  &  lunary-ai/lunary                    & 1.1k                          &    LLM applications                                                                                                  \\ \hline
PortKey                               & Google  &  Portkey-AI/gateway                    & 6.3k                          &    LLM applications                                                                                                   \\ \hline
TraceLoop                             & Google & traceloop/openllmetry                 & 3.4k                          &     LLM applications                                                                                              \\ \hline

Trulens                               & Google  &   truera/trulens                        & 2.2k                          &   LLM applications                                                                                                  \\ \hline

\bottomrule

\end{tabular}
\end{table*}

\subsection{AgentOps-Relevant Tools} \label{subsec:AgentOps-Relevant Tools}

This study includes a diverse range of 17 tools relevant to AgentOps, designed to enable observability in agents or LLM applications. Table \ref{tab:Tool List} highlights the AgentOps relevant tools currently available in the market. 
An analysis of GitHub star rankings reveal that many of these tools have received significant attention, with 14 out of 17 tools accumulating thousands of stars. As of November 2024, several tools stand out as leaders in their respective categories, such as AgentOps (1.7k stars) and AgentNeo (1k stars). Among tools designed for LLM applications, Langfuse (6.5k stars), PortKey (6.3k stars), and Arize (3.9k stars), TraceLoop (3.4k stars), and DataDog (2.9k stars) stand out as leading solutions. It is worth noting that some tools may have been unintentionally excluded due to keyword mismatches or incomplete documentation. We aim to address this limitation by monitoring emerging tools and incorporating them into future work.

\begin{table*}[h]
\centering
\footnotesize
\caption{Key Features of AgentOps relevant Tools}
\label{tab:Key Features Mapping}
\begin{tabular}{l|c|c|c|c|c|c|c}
\toprule \hline
\textbf{Tool} & \textbf{Customisation} & \textbf{Prompt Mgmt} & \textbf{Evaluation} & \textbf{Feedback} & \textbf{Monitoring} & \textbf{Tracing} & \textbf{Guardrails} \\
\hline
Agenta     & Yes & Yes & Yes & Yes & Yes & Yes & No  \\
AgentNeo   & Yes & No  & Yes & No  & Yes & Yes & No  \\
AgentOps   & Yes & Yes & Yes & Yes & Yes & Yes & Yes \\
AGIFlow    & Yes & Yes & Yes & No  & Yes & Yes & No  \\
Arize      & No  & No  & Yes & No  & Yes & Yes & Yes \\
Datadog    & No  & No  & No  & No  & Yes & Yes & No  \\
Dify       & Yes & Yes & Yes & Yes & Yes & Yes & Yes \\
Helicone   & No  & No  & Yes & No  & Yes & Yes & No  \\
Langfuse   & No  & Yes & Yes & Yes & Yes & Yes & No  \\
LangTrace  & No  & No  & Yes & No  & Yes & Yes & No  \\
LangSmith  & No  & Yes & Yes & Yes & Yes & Yes & Yes \\
Lunary     & No  & Yes & Yes & Yes & Yes & Yes & Yes \\
TraceLoop  & No  & No  & No  & No  & Yes & Yes & No  \\
Trulens    & No  & Yes & Yes & Yes & Yes & Yes & Yes \\
Portkey    & No  & No  & Yes & No  & Yes & Yes & No  \\ \hline 
\bottomrule 
\end{tabular}
\end{table*}

\subsection{Key Features} \label{subsec:key features}
% \textcolor{red}{Qinghua: this is study results, not methodology. you can create a new section called mapping study results. mapping the tools with the features you summarised here. make sure the classificatons/structure are aligned with your taxonomy. for example, add context management (e.g. related papers/tools/existing-workflows to scientists, early evaluation results to the original prompts/workflow/plan for optimisation) and workflow/plan.
% see table 1: https://arxiv.org/pdf/2301.11616. you can use the tools or use features as index to map out the analysis}

We summarized the key features of the identified AgentOps relevant tool, as shown in Table \ref{tab:Key Features Mapping} and Table \ref{tab:AgentOps Functions}.

\begin{table*}[h]
\center
\footnotesize
\caption{AgentOps Relevant Tools Key Features}
\label{tab:AgentOps Functions}
\begin{tabular}
{@{\extracolsep\fill}L{20mm}|L{45mm}|L{85mm}}  
\toprule \hline
\textbf{Category}                         & \textbf{Features}                                      & \textbf{Description}                                                                                                                          \\  \hline 
\multirow{7}{15mm}{Customization}              & Provision, custom, spawn, and deploy autonomous agents    & Create customisable and scalable autonomous agents.                                                                                               \\ \cline{2-3} 
                                             & Extend agent capabilities with toolkits                    & Add toolkits from marketplace to agent workflows.                                                                                             \\ \cline{2-3} 
                                             & Extend agent capabilities with multiple vector databases         & Connect to multiple vector databases to improve agent's performance.                                                                                          \\ \cline{2-3} 
                                             & Extend agent capabilities with fine-tuned models         & Custom fine-tuned models for business specific use cases.                                                                                           \\ \hline
\multirow{4}{15mm}{Prompt Management}           & Prompt versioning and management                           & Keep track of different versions of prompts used in agents. Useful for A/B testing and optimizing agent performance.                                        \\ \cline{2-3} 
                                             & Prompt playground with model comparisons                   & Test and compare different prompts and models for agents before deployment.                                                                             \\ \cline{2-3} 
                                             & Prompt injection detection                                 & Identify potential code injection and secret leaks.                                                                                                          \\ \hline
\multirow{6}{15mm}{Evaluation}         & Test agents against benchmarks and leaderboards.      & Create a dataset, define metrics, run evaluations, compare results, track results over time etc. \\ \cline{2-3} 
                                             & \multirow{5}{*}{Evaluate agents in diverse step}       & Evaluate Final Response- Evaluate the agent's final response.                                                                                                \\ \cline{3-3} 
                                             &                                                            & Evaluate Single step-Evaluate any agent step in isolation (\eg whether it selects the appropriate tool).                                                  \\ \cline{3-3} 
                                             &                                                            & Evaluate Trajectory- Evaluate whether the agent took the expected path (\eg of tool calls) to arrive at the final answer.                              \\ \hline

\multirow{3}{15mm}{Feedback} & Collect explicit feedback& Directly prompt the user to give feedback, this can be a thumb up or a thumb down.\\

\cline{2-3}&Collect implicit feedback& Measure the user's behavior, this can be time spent on a page, click-through rate. \\ \hline

\multirow{2}{15mm}{Monitoring}            & Agent analytics  dashboard                                 & Monitor diverse level and dimension statistics metrics about agents.                                                                                  \\ \cline{2-3} 
                                             & LLM cost management and tracking                           & Track spend (token cost) with foundation model providers.                                                                              \\ 
    
                                             \hline
 \multirow{3}{15mm}{Tracing}                                             & \multirow{3}{45mm}{LLM/agent tracing} &                                         Trace each agent span, \eg the whole chain, retrieval, LLM call, tool call etc.                                                                                                                                    \\ \cline{3-3} 
                                             &                                 & Trace evaluation span                                                                                                                \\  \cline{3-3} 
                                             &                                                            & Trace user feedback                                                                                                                                             \\ \hline
 \multirow{3}{15mm}{Guardrails}               & Predefined rules and constraints                           & Set rules or constraints to limit agent actions, ensuring safe and predictable behavior.                                                                                      \\ \cline{2-3} 
                                             & Fallback and escalation paths                              & Provide safe defaults or redirect cases to human operators in ambiguous or risky scenarios.                                                                                   \\ \hline

\bottomrule

\end{tabular}
\end{table*}

\subsubsection{Customization} \label{subsubsec:agent creation feature}

When creating agents, existing tools extend agent capabilities by adding toolkits from marketplaces to agent workflows, connecting to multiple knowledge bases to improve performance, and integrating customized fine-tuned models for specific business use cases. 

\subsubsection{Prompt Management} \label{subsubsec:prompt management feature}
Prompt versioning and management allow developers to store and track different versions of prompts, making it invaluable for testing, optimizing, and reusing prompts across various stages of agent production.
The prompt playground enables developers to edit, import, and test various prompt templates with different models, helping to compare performance before deployment. Additionally, consistent monitoring of prompts is also necessary for maintaining the reliability and security of agents, particularly in detecting and mitigating issues such as code injection attacks and secret leaks embedded within prompts.

\subsubsection{Evaluation} \label{subsubsec:evaluation and testing feature}

Evaluation is the process of assessing the behaviour and capabilities of an agent against specific criteria or general benchmarks.
The typical evaluation process involves creating a suitable evaluation dataset, defining clear criteria and metrics, and conducting thorough testing based on these predefined metrics. It is important to test the agent's performance against user requirements, standard leaderboards or comparable systems. For agents, evaluation goes beyond simply assessing the the final output. It is equally important to monitor and track the agent's execution steps and evaluate intermediate outputs to ensure the entire process meets the intended goals and aligns with governance requirements.

LangSmith\footnote{LangSmith.\url{https://docs.smith.langchain.com/evaluation}} introduces two additional dimensions of agent evaluation: 1) Step-by-Step Evaluation: Assess each individual step the agent takes in isolation, such as determining whether it selects the appropriate tool. 2) Trajectory Evaluation: Examine whether the agent followed the expected sequence of actions, including the series of tool calls, to arrive at the final answer. This ensures that the decision-making process is sound, not just the outcome.

\subsubsection{Feedback}  \label{subsubsec:feedback feature}
Human feedback plays a key role in evaluating the quality of an agent's output.
Feedback is collected as a score and attached to an execution trace or an individual LLM generation. Feedback can be also used to retrain/fine-tune LLM or improve the design of agents (e.g. stored in memory or used in prompts as positive or negative examples).
Langfuse\footnote{Langfuse.\url{https://langfuse.com/docs/scores/user-feedback}} defined different types of feedback that can be collected that vary in quality, detail, and quantity: 1) Explicit Feedback: Directly prompt the user to give feedback, this can be a rating, a like, a dislike, a scale or a comment. While it is simple to implement, quality and quantity of the feedback is often low. More structured and fine-grained feedback is expected. 2) Implicit Feedback: Measure the user's behavior, \eg time spent on a page, click-through rate, accepting or rejecting final output. This type of feedback is more difficult to implement but is often more frequent and reliable. 

\subsubsection{Monitoring} \label{subsubsec:monitoring feature}

Developers can continuously monitor agent performance and enhance observability throughout the agent's execution process by closely tracking its outputs. This involves keeping track of monitoring metrics (\eg latency and cost), associating feedback with agent spans to evaluate performance, and debugging issues by diving into specific traces and spans where errors occurred. Monitoring also helps identify the root causes of unexpected results, errors, or latency issues, allowing developers to optimize performance based on real-time feedback.

\subsubsection{Tracing} \label{subsubsec:tracing feature}
AgentOps is designed to support developers in transitioning from prototype to production, ensuring that the work does not stop once the agent is created and initial tests are passed.
Within the otool, agents execute increasingly complex tasks and iterative runs, such as chains, tool-assisted agents, and advanced prompts. By adding traces, AgentOps captures the entire process—from the moment a user sends a prompt to the final output—helping developers understand each step and identify the root causes of any issues. 
Execution tracing allows developers to follow the agent’s decision-making process step-by-step, providing insights into the flow of actions, decisions, and interactions. This helps identify where errors or unexpected behaviors originate, making it easier to isolate and address problems within complex agentic workflows.
There is no doubt that tracing is the most direct way to enable observability in AgentOps platform. All tools listed in the Table \ref{tab:Tool List} have implemented tracing functionality.

\subsubsection{Guardrails} \label{subsubsec:guardrails feature}
Guardrails for agent application are essential for ensuring AI-safety-by-design. 
Arize have already integrated Arize Guards\footnote{Arize Guards. \url{https://docs.arize.com/arize/llm-monitoring-and-guardrails/guardrails}}. Additionally, an increasing number of tools (\eg AgentNeo\footnote{AgentNeo. \url{https://github.com/raga-ai-hub/agentneo}} ) have Guardrails implementation on their planned feature lists, aiming to enhance the safety of agents. AgentOps tools can trace the activation, execution process and outcomes of guardrails, which can be used to generate safety cases for auditing purposes.

\section{Taxonomy of AgentOps} 
\label{sec:Taxonomy}

\begin{figure*}
\centering
 \includegraphics[width=0.9\textwidth]{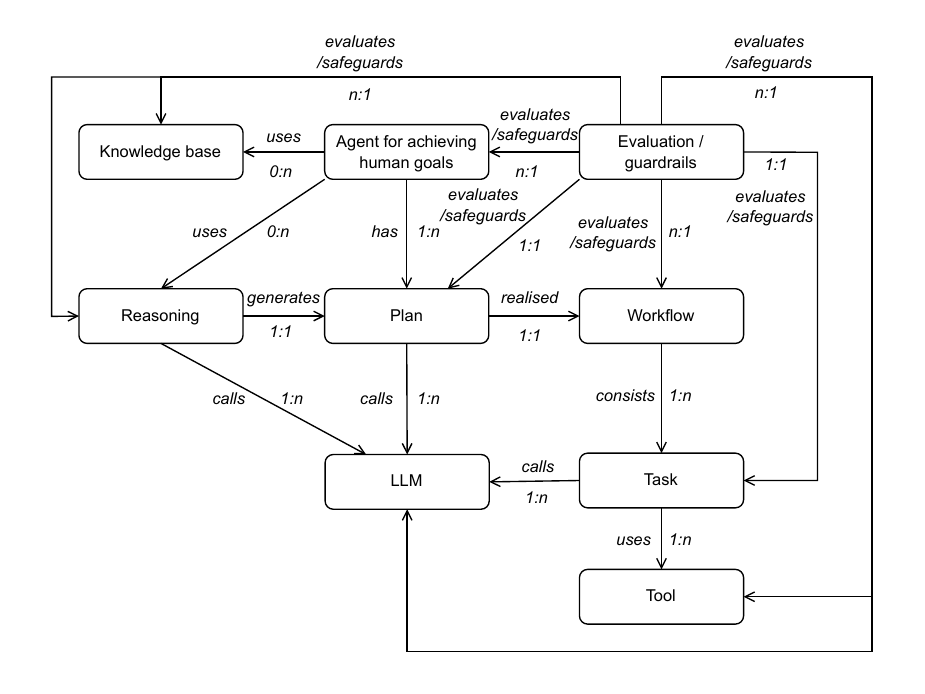}
\caption{Entity-Relationship Model for Agent Artifacts} \label{fig:Relationship}
\end{figure*}

Once agent developer have deployed their agent, they need trace its execution process to monitor agent performance and evolution of agent artifacts.
The AgentOps tool provides tracing functions to troubleshoot performance and correlate data throughout the product, enabling developer to find and resolve issues in agents.

In this section, we first introduce an entity-relationship model to describe various relationships between different agent artifacts. Then, we present a comprehensive taxonomy of Agentops, which serves as a template for developers to design and implement AgentOps tools for tracing agent artifacts and their associate data.

\subsection{Entity-Relationship Model for Agent Artifacts}

To clarify the relationships between traceable artifacts, Figure \ref{fig:Relationship} illustrates the connections among nested spans in an agent trace.

When an \textbf{agent} receives a user \textbf{goal}, it can utilize zero or more \textbf{knowledge bases} to gather information or support decision-making. These knowledge bases supply essential data and contextual information for the agent span.
A single \textbf{reasoning} span generates a \textbf{plan}, structured through logical processes and analysis.

An \textbf{agent} can generate multiple plan spans, each representing a structured plan to address specific objectives. The \textbf{agent} relies on these plans to organize and coordinate its tasks. Each \textbf{plan} span may call one or more \textbf{LLM} spans to leverage the LLM's processing capabilities, aiding in the execution of planned actions. A \textbf{plan} is realized as a single Workflow, translating the strategic plan into practical execution.

A \textbf{workflow} comprises multiple \textbf{tasks}, with each \textbf{task} representing a specific action within the larger \textbf{workflow}. These \textbf{task} collectively fulfill the goals outlined in the \textbf{workflow}. Each \textbf{task} may utilise one or more \textbf{tools}, providing additional functionality or resources essential for task execution. \textbf{Task} may also call one or more \textbf{LLM} spans to access model-based functionalities, such as prediction or analysis, that are critical for task completion.

\textbf{Evaluation} spans assess either a specific \textbf{agent}, a \textbf{plan} or a single \textbf{workflow}, ensuring that the agent’s overall performance or the effectiveness of the workflow aligns with intended goals. The \textbf{guardrail} monitors all other spans in the agent’s lifecycle, enforcing constraints and ensuring compliance with predefined rules. This universal connection helps maintain the safety of the agent’s actions across all spans.

\begin{figure*}
\centering
 \includegraphics[width=0.83\textwidth]{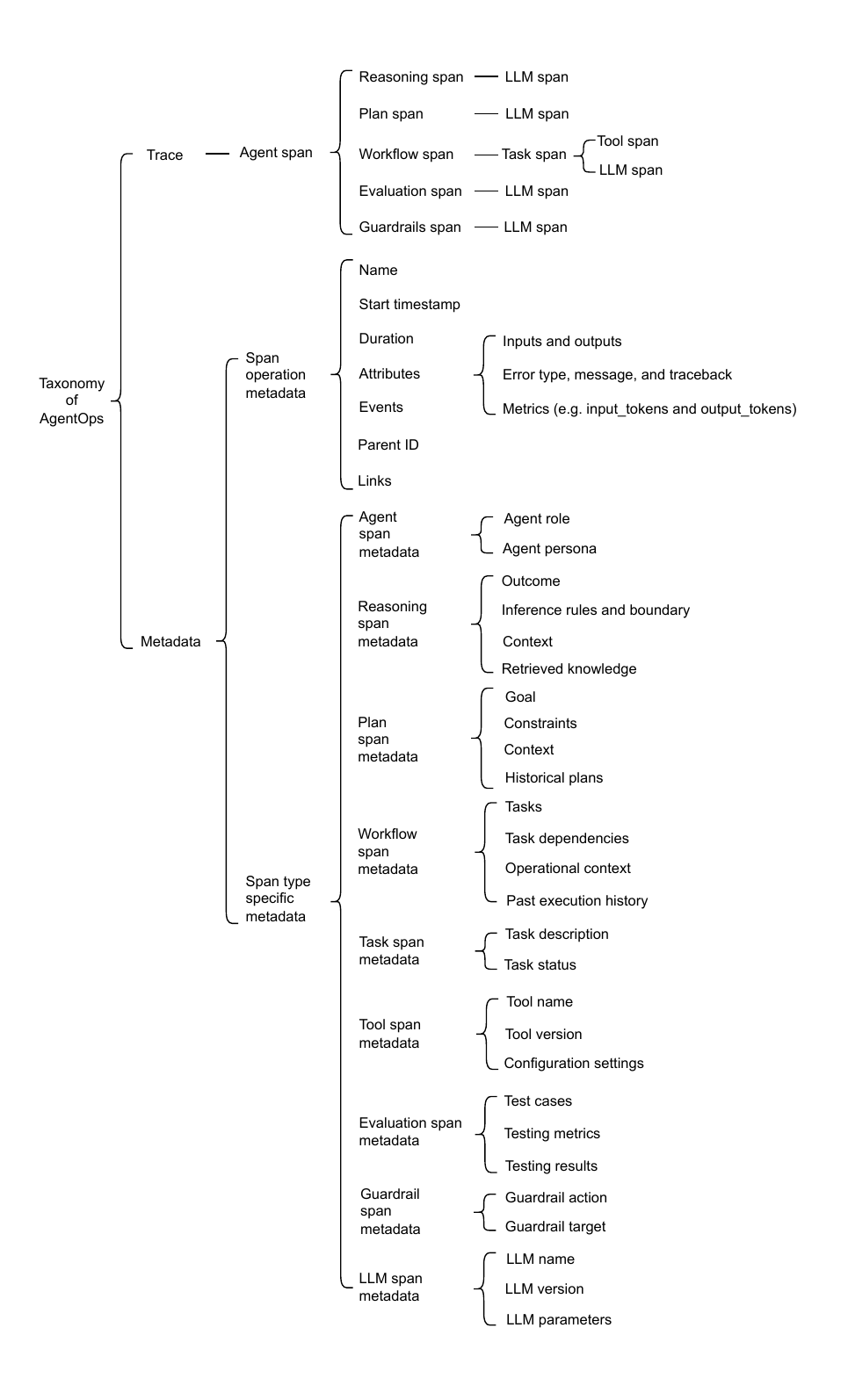}
\caption{Taxonomy of AgentOps} \label{fig:Taxonomy}
\end{figure*}

\subsection{Taxonomy of AgentOps}

In AgentOps, a \textbf{trace} reveals the entire process from the moment a user submits a goal achieving request to the point when the final result is
delivered. This includes the reasoning process, the plan generated, the workflow along with its associated tasks, the knowledge retrieved, the tools invoked, the evaluation and guardrails applied, and multiple LLM calls. 

A \textbf{trace} consists of one or more \textbf{spans}. The first span represents the root span. Each root span represents a request from start to finish. The spans beneath the parent provide deeper context for what occurs during a request, detailing the steps that make up the request.

A\textbf{span} could consists of the following metadata attributes:

\begin{itemize}
    \item \textbf{Name}: The label or identifier of the span, indicating the type of operation being performed.

    \item \textbf{Start Timestamp}: The exact time when the span begins, providing temporal context for tracking and performance analysis.

    \item \textbf{Duration}: The total time taken for the span’s operation, measured from start to finish. This metric helps in analyzing efficiency and identifying potential performance bottlenecks.

    \item \textbf{Attributes}:
    \begin{itemize}
        \item \textbf{Inputs and Outputs}: Data fed into the span (e.g., user goal) and the resulting outputs (e.g., tool calling result or final result).
        \item \textbf{Error Type, Message, and Traceback}: Information about any errors encountered, including the error type, a descriptive message, and traceback details for debugging.
        %\item \textbf{Metadata}: Additional information relevant to the operation, such as LLM parameters (e.g., temperature, max\_tokens) which can impact the behavior of the span.
        \item \textbf{Metrics}: Quantitative data related to the span, such as input\_tokens, output\_tokens, evaluation metrics, monitoring metrics, which measure and track cost usage and agent performance.
    \end{itemize}

    \item \textbf{Events}: Specific occurrences or a detailed timeline of actions and transitions within each span, providing detailed insights into the span's activities or any notable events during its lifecycle.

    \item \textbf{Parent ID}: An identifier that links this span to its parent span, establishing a hierarchical relationship between spans and helping to track nested operations.

    \item \textbf{Links}: Connections to other spans or external references, which can help in understanding dependencies and relationships between different spans in a complex workflow.
\end{itemize}

In the context of agent spans, where a trace consists of a series of nested spans representing different operational spans or runs, each type of span involved in agent span plays a distinct role. The agent span includes the following nested spans:

\begin{itemize} 
\item \textbf{Agent Span:} The agent span metadata includes the agent's role and persona, which significantly influence how the agent performs tasks, interacts with users and makes decisions. The agent span meta data includes the following.
  \begin{itemize}
      \item \textbf{Agent Role:} The scope or responsibility of the agent.
      \item \textbf{Agent Persona:} The behavioral characteristics and interaction style the agent adopts.
  \end{itemize}

\item  \textbf{Reasoning Span:} Reasoning span, captures the reasoning processes of the agent. Reasoning Span metadata including:
\begin{itemize}
\item\textbf{Context:} The relevant information or situational data that informs the reasoning process. This may include inputs from previous spans or external factors that influence the agent's reasoning.

\item\textbf{Retrieved Knowledge:} Information or data that the agent retrieves and references during reasoning, possibly from external sources or memory, to support its reasoning.

\item\textbf{Inference Rules and Boundary:} The logical rules, constraints, or boundaries applied during the reasoning. This could include specific guidelines or conditions under which the agent performs reasoning.

\item\textbf{Outcome:} The thoughts generated or conclusion reached after the reasoning process.
\end{itemize}

\item \textbf{Planning Span:} Planning span records the planning phase, where the agent outlines the steps or objectives required to achieve the goal. It defines the intended sequence of operations, setting up a structured path for the agent's activities. Planning Span metadata including:

\begin{itemize}
\item
\textbf{Goal:} The specific objective or desired outcome that the agent intends to achieve through this plan. 
\item
\textbf{Constraints:} The limitations or restrictions within which the plan must operate. These constraints can include time limits, resource limitations, or predefined rules that shape the planning process.
\item
\textbf{Context:} Relevant situational or environmental information that informs the plan. 
\item
\textbf{Historical Plans:} Records of previous plans that may influence the current planning process. This includes past strategies or actions taken in similar contexts, which can provide valuable insights or best practices for the current plan.
\end{itemize}

\item \textbf{Workflow Span:} Within the workflow span, the trace documents the organization of tasks through task spans and tool Spans. The workflow span, supported by an LLMspan, details how individual tasks are broken down and managed. The task span represents specific actions within the workflow, while the tool span (nested within the task span) logs interactions with tools or external resources the agent uses to fulfill specific parts of the workflow.

The metadata of the workflow span includes the following.

\begin{itemize}
\item
\textbf{Tasks:} A list of task span that need to be completed as part of the workflow. This defines the specific actions or steps the agent needs to perform within this span.
\item\textbf{Task Dependencies:} Information about dependencies between tasks, indicating the order in which tasks should be executed or any prerequisites required for specific tasks. This helps manage the sequence and ensures that tasks are executed in a logical, efficient manner.
\item\textbf{Operational Context:} The situational information or environment details relevant to executing the workflow. This can include real-time conditions, status updates from other spans, or external factors that might influence task execution.
\item\textbf{Past Execution History:} Records of previous executions of similar workflows or tasks in (long-term) memory module, which can provide insights into best practices, potential pitfalls, or optimisation opportunities for the current workflow.
\end{itemize}

\item  \textbf{Task Span:} The task span represents a discrete unit of work or action within a workflow. It is a fundamental part of the workflow structure, defining individual tasks that the agent needs to execute in a sequence or parallel arrangement. The task span metadata includes:
    \begin{itemize}
        \item \textbf{Task Description:} Specific information about the task to be performed, including task objectives, instructions, and parameters needed for execution.
        
        \item \textbf{Task Status:} The current status (e.g., pending, in progress, completed) and the result of the task, which could include success, failure, or a specific output generated by the task.
    \end{itemize}

     \item  \textbf{Tool Span:} The tool span represents interactions with external tools or resources that assist in task execution. This span captures details about tools utilized by the agent, logging their configurations, responses, and any intermediate outputs. Tool Span metadata includes:
    \begin{itemize}
        \item \textbf{Tool Name:} The name of the tool.

        \item \textbf{Tool Version:} The version of the tool.
        
        \item \textbf{Configuration Settings:} Parameters or settings configured for the tool during its use, such as version restriction, input formats, timeouts, or resource limits that may influence its behavior.
    
    \end{itemize}

\item
\textbf{Evaluation Span:} 
The evaluation span assesses the correctness and quality of the agent's actions and outputs performance against predefined criteria. It verifies whether the agent's actions align with expectations or quality standards, providing a feedback mechanism to ensure that outputs meet the intended goals.

The metadata of the evaluation span includes the following.

\begin{itemize}
\item \textbf{Test Cases:} Specific scenarios or conditions under which the agent’s performance or outputs are evaluated. These cases provide a structured way to assess the agent’s actions and outcomes against expected behavior~\cite{xia2024evaluationdrivenapproachdesigningllm}.

\item\textbf{Testing Metrics:} Quantitative or qualitative measures used to assess the agent’s performance. These metrics could include accuracy, efficiency, relevance, or other criteria that define the quality of the agent's output.
\item\textbf{Testing Results:} The actual outcomes or findings from the evaluation process, indicating how well the agent's performance aligns with the expected standards defined in the test cases and metrics.
\end{itemize}
    
\item  \textbf{Guardrail Span:} The guardrail span defines the guardrails applied to ensure the agent’s operations are aligned with expected governance requirements~\cite{md2024guardrails}. It helps prevent errors or unintended actions by setting boundaries on the agent’s behavior. Guardrail span metadata includes:
    \begin{itemize}

        \item \textbf{Guardrail Actions:} Guardrails actions triggered, such as block, validation, filter, etc.

        \item \textbf{Guardrail Targets:} Specific agent artifacts that guardrails are applied, such as goals and tools.
    \end{itemize}

\item  \textbf{LLM Span:} The LLM Span captures interactions with an LLM, where the model processes language-based inputs to generate responses or insights. This span is essential for tasks involving natural language understanding, generation, or interpretation. LLM Span metadata could includes:
    \begin{itemize}
        \item \textbf{LLM Name:} The name of the LLM.
        \item \textbf{LLM Version:} The version of the LLM.

        \item \textbf{LLM Parameters:} Settings applied to the LLM, such as temperature (affecting randomness), max\_tokens (limiting response length), and other relevant hyper-parameters that shape the output.
        
    \end{itemize}

\end{itemize}

%\section{Discussion} \label{sec:discussion}
%\textcolor{blue}{
%(Liming Dong) In progress...}
%\subsection{Key findings}
%\subsection{Limitation}
%\subsection{Future work/roadmap}

%\section{Related Work} \label{sec:related work}
%\textcolor{blue}{
%(Liming Dong) In progress...}

%\subsection{Importance of Agent System Reliability/Observability}

%\subsection{Practices of Agent System Observability/Visibility}
%1.TapeAgents: a Holistic Framework for Agent Development and Optimization\\
%2.Visibility into AI Agents\\
%3.AgentLens: Visual Analysis for Agent Behaviors
%in LLM-based Autonomous Systems\\
%4.The Dawn of GUI Agent: A Preliminary Case Study with Claude 3.5 Computer Use\\
%...

\section{Threats to Validity} \label{sec:ttv}

\textbf{Tool Selection Limitations:} Due to the rapid proliferation of various tools and AI platforms, it is possible that not all relevant AgentOps tools were identified. To address this limitation, we selected tools from multiple data sources. The identified tools include both open-source AgentOps tools, such as AgentOps and Langfuse, as well as commercial observability platforms like Datadog.

\noindent\textbf{Data Coverage Limitations:} The comprehensive overview of traceable artifacts throughout the AgentOps life-cycle provided in this work may not encompass all possible data attributes related to AI agents.
To ensure broader coverage of important traceable data across the entire life-cycle of an agent and enrich on the data attributes outlined in our paper, we have drawn on some relevant academic literature  \cite{PromptSurvey,md2024guardrails,chan2024visibility} to support our findings as well.
However, some potentially valuable data, such as trace links and interactions between different steps, may have been missed. Future work will aim to explore these gaps further.

\section{Conclusion} \label{sec:conlusion}

In this paper, we proposed AgentOps, a specialised DevOps paradigm tailored for LLM agents to enable observability. Through a systematic mapping study of existing AgentOps relevant tools, we first proposed an entity-relationship model to understand the key agent artifacts and their relationship. Then we presented a comprehensive taxonomy of AgentOps, offering a structured template for developers to design and implement AgentOps tools. Future research will focus on validating the proposed taxonomy through real-world case studies and the development of a AgentOps tool prototype.

\bibliographystyle{IEEEtran}
\bibliography{main}

% Generated by IEEEtran.bst, version: 1.14 (2015/08/26)
\begin{thebibliography}{10}
\providecommand{\url}[1]{#1}
\csname url@samestyle\endcsname
\providecommand{\newblock}{\relax}
\providecommand{\bibinfo}[2]{#2}
\providecommand{\BIBentrySTDinterwordspacing}{\spaceskip=0pt\relax}
\providecommand{\BIBentryALTinterwordstretchfactor}{4}
\providecommand{\BIBentryALTinterwordspacing}{\spaceskip=\fontdimen2\font plus
\BIBentryALTinterwordstretchfactor\fontdimen3\font minus \fontdimen4\font\relax}
\providecommand{\BIBforeignlanguage}[2]{{%
\expandafter\ifx\csname l@#1\endcsname\relax
\typeout{** WARNING: IEEEtran.bst: No hyphenation pattern has been}%
\typeout{** loaded for the language `#1'. Using the pattern for}%
\typeout{** the default language instead.}%
\else
\language=\csname l@#1\endcsname
\fi
#2}}
\providecommand{\BIBdecl}{\relax}
\BIBdecl

\bibitem{bommasani2021opportunities}
R.~Bommasani, D.~A. Hudson, E.~Adeli, R.~Altman, S.~Arora, S.~von Arx, M.~S. Bernstein, J.~Bohg, A.~Bosselut, E.~Brunskill \emph{et~al.}, ``On the opportunities and risks of foundation models,'' \emph{arXiv preprint arXiv:2108.07258}, 2021.

\bibitem{liu2024agentdesignpatterncatalogue}
\BIBentryALTinterwordspacing
Y.~Liu, S.~K. Lo, Q.~Lu, L.~Zhu, D.~Zhao, X.~Xu, S.~Harrer, and J.~Whittle, ``Agent design pattern catalogue: A collection of architectural patterns for foundation model based agents,'' 2024. [Online]. Available: \url{https://arxiv.org/abs/2405.10467}
\BIBentrySTDinterwordspacing

\bibitem{lu2024towards}
Q.~Lu, L.~Zhu, X.~Xu, Z.~Xing, S.~Harrer, and J.~Whittle, ``Towards responsible generative ai: A reference architecture for designing foundation model based agents,'' in \emph{2024 IEEE 21st International Conference on Software Architecture Companion (ICSA-C)}.\hskip 1em plus 0.5em minus 0.4em\relax IEEE, 2024, pp. 119--126.

\bibitem{lu2023responsible}
Q.~Lu, L.~Zhu, J.~Whittle, X.~Xu \emph{et~al.}, \emph{Responsible AI: Best Practices for Creating Trustworthy AI Systems}.\hskip 1em plus 0.5em minus 0.4em\relax Addison-Wesley, 2023.

\bibitem{bass25engineering}
L.~Bass, Q.~Lu, I.~Weber, and L.~Zhu, \emph{Engineering AI Systems: Architecture and DevOps Essentials}.\hskip 1em plus 0.5em minus 0.4em\relax Addison-Wesley, 2025.

\bibitem{keele2007guidelines}
B.~Kitchenham and S.~Charters, ``Guidelines for performing systematic literature reviews in software engineering version 2.3,'' Software Engineering Group, School of Computer Science and Mathematics, Keele University and Department of Computer Science University of Durham, Tech. Rep., 2007.

\bibitem{MultiSLR2019}
\BIBentryALTinterwordspacing
V.~Garousi, M.~Felderer, and M.~V. M{\"{a}}ntyl{\"{a}}, ``Guidelines for including grey literature and conducting multivocal literature reviews in software engineering,'' \emph{Inf. Softw. Technol.}, vol. 106, pp. 101--121, 2019. [Online]. Available: \url{https://doi.org/10.1016/j.infsof.2018.09.006}
\BIBentrySTDinterwordspacing

\bibitem{xia2024evaluationdrivenapproachdesigningllm}
\BIBentryALTinterwordspacing
B.~Xia, Q.~Lu, L.~Zhu, Z.~Xing, D.~Zhao, and H.~Zhang, ``An evaluation-driven approach to designing llm agents: Process and architecture,'' 2024. [Online]. Available: \url{https://arxiv.org/abs/2411.13768}
\BIBentrySTDinterwordspacing

\bibitem{md2024guardrails}
\BIBentryALTinterwordspacing
M.~Shamsujjoha, Q.~Lu, D.~Zhao, and L.~Zhu, ``Designing multi-layered runtime guardrails for foundation model based agents: Swiss cheese model for ai safety by design,'' 2024. [Online]. Available: \url{https://arxiv.org/abs/2408.02205}
\BIBentrySTDinterwordspacing

\bibitem{PromptSurvey}
\BIBentryALTinterwordspacing
S.~Schulhoff, M.~Ilie, N.~Balepur, K.~Kahadze, A.~Liu, C.~Si, Y.~Li, A.~Gupta, H.~Han, S.~Schulhoff, P.~S. Dulepet, S.~Vidyadhara, D.~Ki, S.~Agrawal, C.~Pham, G.~Kroiz, F.~Li, H.~Tao, A.~Srivastava, H.~D. Costa, S.~Gupta, M.~L. Rogers, I.~Goncearenco, G.~Sarli, I.~Galynker, D.~Peskoff, M.~Carpuat, J.~White, S.~Anadkat, A.~Hoyle, and P.~Resnik, ``The prompt report: A systematic survey of prompting techniques,'' 2024. [Online]. Available: \url{https://arxiv.org/abs/2406.06608}
\BIBentrySTDinterwordspacing

\bibitem{chan2024visibility}
A.~Chan, C.~Ezell, M.~Kaufmann, K.~Wei, L.~Hammond, H.~Bradley, E.~Bluemke, N.~Rajkumar, D.~Krueger, N.~Kolt \emph{et~al.}, ``Visibility into ai agents,'' in \emph{The 2024 ACM Conference on Fairness, Accountability, and Transparency}, 2024, pp. 958--973.

\end{thebibliography}

\end{document}